\documentclass[letter, 10pt, conference]{ieeeconf}
\IEEEoverridecommandlockouts                              
\usepackage{cite}
\usepackage{booktabs}
\usepackage{makecell}
\usepackage{algorithm}
\usepackage{algpseudocode}
\usepackage{amsmath}
\usepackage{amsfonts}
\usepackage{amssymb}  
\usepackage[utf8]{inputenc}
\usepackage[english]{babel}
\usepackage[T1]{fontenc}
\usepackage{url}
\usepackage{soul}
\usepackage{color}
\usepackage{graphicx}
\usepackage{xcolor}
\usepackage[export]{adjustbox}
\usepackage{subcaption}

\newcommand{\ours}{\textit{BETR-XP-LLM}}
\begin{document}
\bstctlcite{IEEEexample:BSTcontrol}


\renewcommand\theadfont{\bfseries}
\renewcommand\theadgape{\Gape[4pt]}

\author{Jonathan Styrud\authorrefmark{1}\authorrefmark{3}, Matteo Iovino\authorrefmark{2}, Mikael Norrlöf\authorrefmark{1}, Mårten Björkman\authorrefmark{3} and Christian Smith\authorrefmark{3}
\thanks{The real robot experiments were carried out in the WASP Research Arena (WARA)-Robotics, hosted by ABB Corporate Research Center in Västerås, Sweden. This project is supported by the Wallenberg AI, Autonomous Systems, and Software Program (WASP) funded by the Knut and Alice Wallenberg Foundation. The authors gratefully acknowledge this support.}
\thanks{\authorrefmark{1}ABB Robotics, Västerås, Sweden}
\thanks{\authorrefmark{2}ABB Corporate Research, Västerås, Sweden}
\thanks{\authorrefmark{3}Division of Robotics, Perception and Learning, Royal Institute of Technology (KTH), Stockholm, Sweden}}

\title{\LARGE Automatic Behavior Tree Expansion with LLMs for Robotic Manipulation}

\maketitle

\begin{abstract}
Robotic systems for manipulation tasks are increasingly expected to be easy to configure for new tasks or unpredictable environments, while keeping a transparent policy that is readable and verifiable by humans. We propose the method BEhavior TRee eXPansion with Large Language Models (\ours) to dynamically and automatically expand and configure Behavior Trees as policies for robot control. The method utilizes an LLM to resolve errors outside the task planner's capabilities, both during planning and execution. We show that the method is able to solve a variety of tasks and failures and permanently update the policy to handle similar problems in the future.
\end{abstract}

\section{Introduction}
Modern robots have the capability of solving complex tasks in controlled environments with high reliability and precision. Traditionally, industrial robots have been tasked with large batches, repeating the same program for years. As robots are now entering smaller businesses, the trends are towards ever smaller batches and frequent updates of robot programs. An increasing number of robots are also working in shared workspaces which create more unpredictable environments~\cite{worldrobotics}. For these reasons, the ability to create robot programs/policies quickly without the need for trained programmers and for those programs to be reactive to their environment is becoming increasingly important. Another decisive factor, especially in industrial settings, is that the program must be transparent and readable to enable analysis, editing, and validation.
A growing and popular alternative in robotics that fulfills all these requirements is to represent the policy with Behavior Trees (BTs)~\cite{iovino_survey_2022, colledanchise_behavior_2018}. Other major advantages of using BTs are explicit support for task hierarchy, action sequencing, and inherent modularity. A currently very active research direction is facilitating the creation of BTs with less effort~\cite{iovino_survey_2022, colledanchise_synthesis_2017, colledanchise_learning_2019, colledanchise_towards_2019, rovida_extended_2017, iovino_learning_2021,  styrud_combining_2022, llmbt, llmobtea}. Another, even more active, area is the creation of robot programs of all types with less effort through the use of Large Language Models (LLMs)~\cite{radford2019language, brown2020language, wang2024survey}.

\begin{figure}[tpb!]
    \setlength{\fboxrule}{0pt}
		\framebox{\parbox{3in}{
            \centering
            \includegraphics[width=0.48\textwidth]{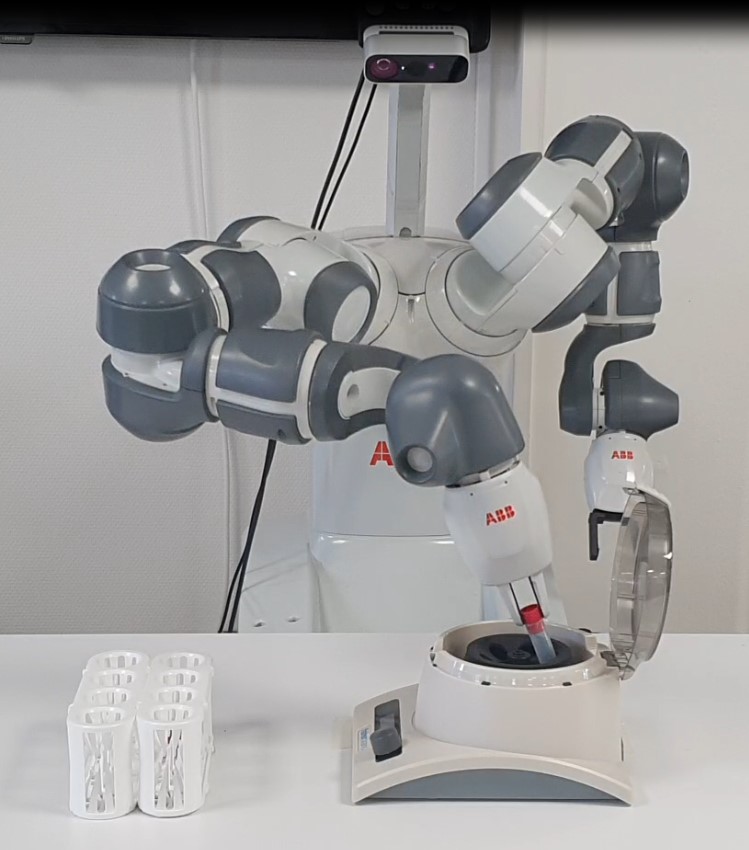}
        }
    }
\caption{ABB YuMi robot performing task 4: inserting a test tube into the centrifuge. The Kinect camera used for object detection can be seen mounted at the top.}
\vspace{-0.4cm}
\label{fig:centrifuge}
\end{figure}

In this context, we present BEhavior TRee eXPansion with Large Language Models (\ours), a method that combines LLMs and task planning to generate a reactive policy in the form of BTs from natural language input. Compared to methods that only generate plans, like SayCan~\cite{saycan}, this minimizes the number of calls to the LLM, saving time and cost. It also makes the policy transparent and verifiable. The overall idea of combining LLMs and planners to create BTs has received some initial exploration recently, most notably in the methods LLM-BT~\cite{llmbt} and LLM-OBTEA~\cite{llmobtea}. \ours~improves on these methods with the following main contributions:
\begin{itemize}
    \item We utilize an LLM beyond goal interpretation to resolve errors outside the planner's capabilities during planning and execution.
    \item We use the LLM output to automatically and permanently update the BT policy, increasing the success rate and robustness, and keeping important properties like transparency and readability while minimizing the number of LLM calls and the amount of manual intervention. An example is shown in Fig.~\ref{fig:bt}.
    \item We show that an improved prompt and LLM compared to LLM-OBTEA~\cite{llmobtea} eliminates the need for reflective feedback and yields better results with fewer LLM calls.
\end{itemize}

\section{Background and Related Work}
We provide relevant background on BTs and research using LLMs to create BTs and other policy representations.

\subsection{LLMs for robot program generation}
In the last few years, the use of Large Language Models (LLMs)~\cite{radford2019language, brown2020language} has grown rapidly, and the field of robotics is not exempt~\cite{wang2024survey}. One of the most prominent examples is SayCan~\cite{saycan} where an LLM interprets an instruction and selects which skill to execute with the help of an affordance score. This has later been extended by others to gain  planning capabilities~\cite{rana2023sayplan, hazra2024saycanpay}. Text2Reaction~\cite{yang2024text2reaction} prompts the LLM again upon discovering problems to resolve them but does not make use of formal planning. 
\par LLMs have also been shown to be able to produce PDDL  format~\cite{aeronautiques1998pddl} problem descriptions from natural language instructions~\cite{liu2023llm+}, which enables the use of PDDL planners to solve the task explicitly.

\begin{figure*}[t!!]
    \vspace{0.1cm}
    \setlength{\fboxrule}{0pt}
		\framebox{\parbox{3in}{

            \hspace*{-0.cm}
                \includegraphics[width=1.08\textwidth]{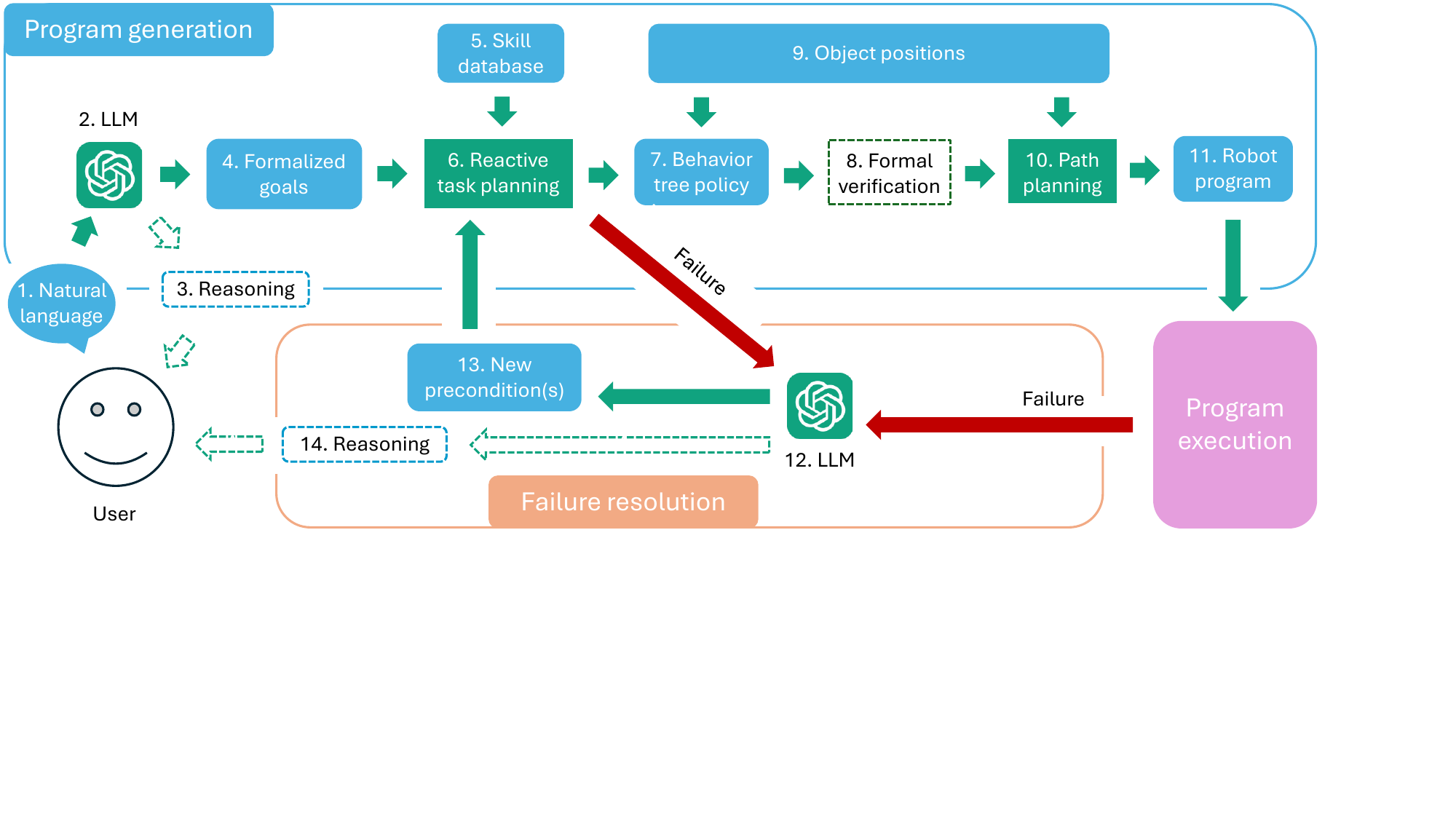}
            }
        }
\vspace{-3.6cm}
\caption{Graphic representation showing all the components of \ours. Green boxes denote algorithms and blue boxes denote data. Dashed lines and boxes denote optional components.}
\vspace{-.0cm}
\label{fig:method}
\end{figure*}

\subsection{Behavior Trees}
Program representations in the works mentioned in the previous section lack the property of being reactive to changes in the environment while still retaining a transparent, readable, verifiable policy. A popular policy representation that does fulfill all those properties is Behavior Trees (BTs). BTs first made debut in computer games but are seeing increasing use in other fields, particularly robotics~\cite{colledanchise_behavior_2018, iovino_survey_2022, hallen2024behavior}.
A BT is a directed tree where a tick signal propagates from the root node to the leaves. Each node runs only if it receives the tick signal and then returns one of the states \textit{Success}, \textit{Failure}, or \textit{Running}. Non-leaf nodes are called \emph{control flow nodes}. The most commonly used control flow nodes are \emph{Sequence}, which ticks children sequentially from left to right, returning once all succeed or one fails, and its counterpart \emph{Fallback} (or \emph{Selector}) which returns when one child succeeds or all fail. Leaves are called \emph{execution nodes} or \emph{behaviors} and are typically separated into the types \textit{Action}(``!'') and \textit{Condition}(``?''). Conditions represent status checks and sensory readings, only returning \textit{Success} or \textit{Failure} while actions represent skills that can take more than one tick to complete and therefore can also return \textit{Running}. Fig.~\ref{fig:bt} shows  BTs for a cube pick and place task.

\subsection{Behavior Tree creation}
Finding more efficient methods to create or generate BTs has received significant research interest in recent years~\cite{iovino_survey_2022}, using learning methods~\cite{colledanchise_learning_2019, iovino_learning_2021, iovino2023framework, mayr22priors}, analytical planners~\cite{tumova_maximally_2014, colledanchise_synthesis_2017, colledanchise_towards_2019, holzl_reasoning_2015, rovida_extended_2017}, improved user interfaces~\cite{gustavsson_combining_2022, gugliermo2023learning, iovino2022interactive}, or various combinations of the three methods~\cite{mayr2022combining, mayr2022skill, styrud_combining_2022, styrud2024bebop}. Recently, methods involving LLMs have also been proposed to interpret the users natural language inputs. Some use an LLM to directly create XML-files defining the BTs~\cite{izzo2024btgenbot, mower2024ros}. These methods do not make use of formal planners, and therefore struggle with tasks requiring long-horizon planning. Another method, LLM-BT~\cite{llmbt}, uses an extensive hard coded parser to translate the LLM's responses into BT nodes and a planner akin to~\cite{colledanchise_towards_2019} to build the BT. LLM-OBTEA~\cite{llmobtea} is very similar but has stricter prompting to obtain the goal conditions directly with minimal parsing. LLM-OBTEA does however use an additional step called reflective feedback where detected syntactical errors are iteratively fed to the LLM in several calls until a response without errors is received. Its extension HOBTEA~\cite{chen2024efficient} improves the speed of the planning algorithm using an LLM to suggest prioritized areas of the search space. At this time however, an LLM call typically takes orders of magnitude longer than the planning algorithm so its utility is limited.
\par Another notable system is MOSAIC~\cite{wang2024mosaic} which uses a BT structure to guide the repeated prompting of an LLM that performs the task planning. A final combination of LLMs and BTs worth mentioning is~\cite{tagliamonte2024generalizable} where the purpose is to generate explanations and answer the users' questions.
\par To summarize, current methods generate BTs from natural language input with some accuracy, but still require substantial effort in providing knowledge to the planning algorithms.

\begin{table*}[htbp]
\centering
\caption{Comparison of properties of different methods for generating robot programs with the use of LLMs.}
\vspace{-.2cm}
\begin{center}
\begin{tabular}{c c c c c}

\fontsize{8}{0}\selectfont \thead{Methods} & \fontsize{8}{0}\selectfont \thead{Formal long-horizon \\ planning} & \fontsize{8}{0}\selectfont \thead{Reactive \\ policy} & \fontsize{8}{0}\selectfont \thead{Failure \\ resolution} & \fontsize{8}{0}\selectfont \thead{Failure resolution \\ permanently improves policy}\cr
 
\cmidrule(r){1-1} \cmidrule(l){2-5}
SayCan~\cite{saycan} & No & No & No & - \cr
\cmidrule(lr){1-1} \cmidrule(lr){2-5}
LLM+P~\cite{liu2023llm+} & Yes & No & No & - \cr
\cmidrule(lr){1-1} \cmidrule(lr){2-5}
Text2Reaction~\cite{yang2024text2reaction} & No & No & Yes & - \cr
\cmidrule(lr){1-1} \cmidrule(lr){2-5}
LLM-BT~\cite{llmbt}, LLM-OBTEA~\cite{llmobtea} & Yes & Yes & No & No \cr
\cmidrule(lr){1-1} \cmidrule(lr){2-5}
\ours & Yes & Yes & Yes & Yes\cr

\bottomrule
\end{tabular}
\end{center}
\vspace{-.3cm}
\label{Method comparison}
\end{table*}

\section{Method}
Our method \ours~as shown in Fig.~\ref{fig:method} can be divided into two main parts, the sequential program generation and the failure resolution. The underlying code, datasets and full prompts can all be found on our github\footnote{https://github.com/jstyrud/\ours}.
\par The overall program generation follows a similar approach to LLM-BT~\cite{llmbt} and especially LLM-OBTEA~\cite{llmobtea} but has an improved prompt that makes reflective feedback superfluous. In our system, the time spent ticking nodes is negligible and thus not a concern. Therefore we do not use compaction as it hides logic inside the tree, reducing readability with the only benefit being fewer nodes ticked. We also do not use the planning speedup of HOBTEA~\cite{chen2024efficient} as running the planner is almost instantaneous compared to the LLM.
\par Planners can, in theory, solve all common robotics task planning problems given sufficient knowledge in advance. In practice, the engineering effort to provide all that knowledge is substantial, and there are always situations where at least some information is missing. As an example, consider a scenario where a human coworker enters the workspace of a robot and puts a coffee mug right where the robot was about to move. This scenario is highly unlikely to have been part of the original task planning, but must still be dealt with, preferably without human intervention. Therefore, the second main part improves on existing methods for generating BTs from natural language input by handling scenarios where the planner is missing crucial knowledge to solve the task. Upon failure, either during planning or execution, the system identifies the failing action of the current policy and uses an \textit{LLM} as a common sense model to suggest what could be necessary for the action to execute successfully. Not needing manual input, this increases robustness without added time and cost. Overall, the weakest link of \ours~is the capability of the LLM, as the planner will always solve the task if it is given correct information.
To further detail of our method, we go through all the sub-components one by one.
\par\textbf{1. Natural language} input in the form of text without any restriction in format. In one of our experiments we also integrate a speech-to-text translation layer. It is fed to
\par\textbf{2. LLM for goal interpretation}. In our experiments we use GPT4-1106 but any LLM of similar capability will work. Besides the natural language instruction input, the prompt contains a list of objects in the scene, formal goal condition alternatives with descriptions, examples, a short scene description, and strict specifications for the output. The scenario description in most experiments is given by simply providing the values of various conditions in text form, for example ``<mug> is on <table>. <coffee> is in <mug>''. Alternatively, the scenario description could also be obtained by prompting a vision language model~\cite{gao2024physically} as in our real experiments. An optional output is the
\par\textbf{3. and 14. Reasoning} reasoning behind the answer, extracted with text parsing since the response follows a strict format. This is not necessary for the method itself but can be used as feedback to the user or for debugging purposes. The main output however is
\par\textbf{4. Formalized goals} in the form of BT  nodes as one of two components required by the planner together with
\par
\begin{figure}
    \setlength{\fboxrule}{0pt}
		\framebox{\parbox{3in}{
            \centering
\vspace{-0.0cm}
\tabskip=0pt
\valign{#\cr
  \noalign{\hfill}
  \hbox{%
    \begin{subfigure}{0.48\textwidth}
    \centering
    \hspace*{1.64cm}
    \includegraphics[width=1.48\textwidth]{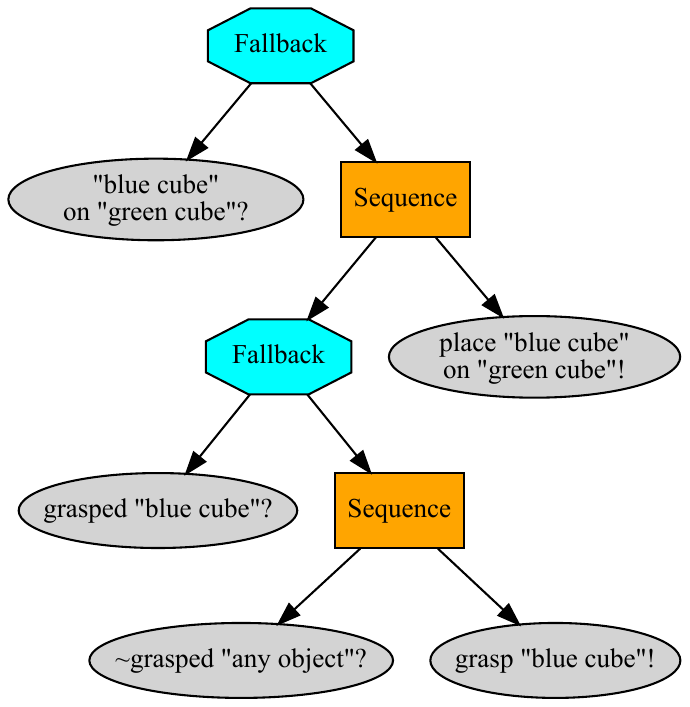}
    \vspace{-4.0cm}
    \caption{Before failure resolution}
    \end{subfigure}%
  }\vspace{.4cm}
  \hbox{%
    \begin{subfigure}{0.48\textwidth}
    \centering
    \includegraphics[width=1.48\textwidth]{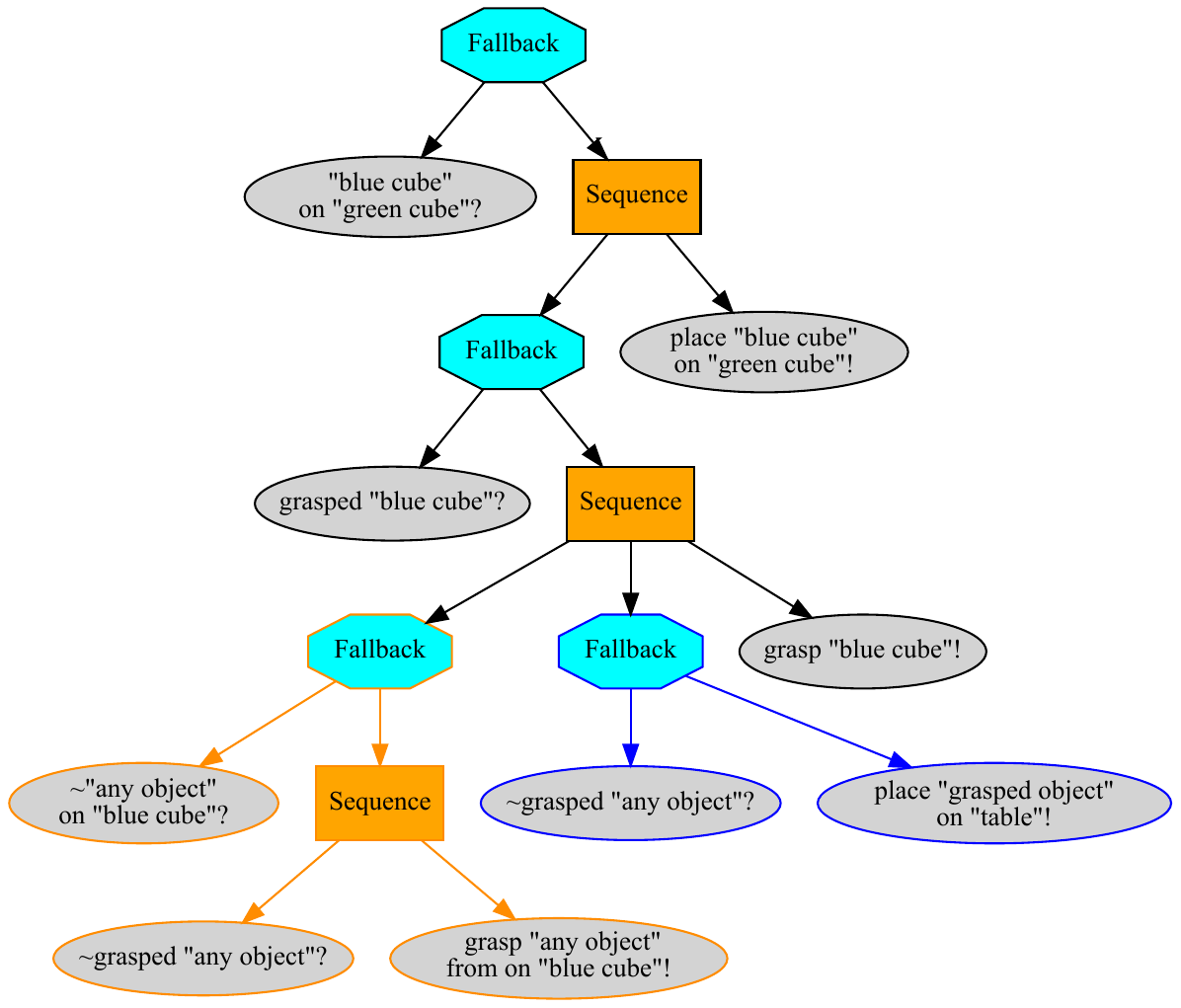}
    \vspace{-1.8cm}
    \caption{After failure resolution}
    \end{subfigure}%
  }\cr
}
}
}
\caption{Example Behavior tree before (a) and after the failure resolution algorithm (b) for a cube pick and place task.}
\label{fig:bt}
\vspace{-0.4cm}
\end{figure}

\textbf{5. Skill database}, a set of parameterized skills that can solve the task. In our experiments they are manually designed but they could also be learned, partially or completely, see for example~\cite{mayr2022skill, styrud2024bebop, yang2018learning,li2019robot}. Example skills are \textit{Grasp} and \textit{Place}, see Fig.~\ref{fig:bt}. For the 
\par\textbf{6. Reactive task planner} to work, the skills must have known preconditions and effects.
In our experiments we use a PDDL style planner adapted from~\cite{colledanchise_towards_2019} that was later extended in~\cite{styrud_combining_2022, gustavsson_combining_2022, iovino2023framework, styrud2024bebop}. The task planning can occur within a simulation, thereby discovering possible errors before the policy is executed on a real system. Which errors are detectable will depend on the completeness of the simulation. The output is a
\par\textbf{7. Behavior tree policy}, deciding which skill to execute depending on the world state. It can be analyzed with
\par\textbf{8. Formal verification} as an optional step. This is made possible by the transparency of the behavior tree policy, where an automated or manual analysis ensures the policy's correctness for specific situations~\cite{biggar2020framework}. This could be mandatory in applications where the reliability of the policy is critical due to safety concerns or other reasons. A necessary input to the policy is
\par\textbf{9. Object positions}. For the policy to be reactive it needs an estimate of the world state. This would typically be given by system sensors, mainly vision. The object positions are also provided to the
\par\textbf{10. Path planning} algorithm.
The BT policy only controls which behaviors to execute. The detailed path would be created by a path planning algorithm called from within the respective behaviors and the output becomes the finalized
\par\textbf{11. Robot program} in the form of code or a sequence of references. This is then continuously uploaded/streamed to the robot system. Upon failure, the system moves to
\par\textbf{12. LLM for failure resolution}. Failures can be detected either during the simulations in the reactive task planning phase or during program execution on the real system. Regardless, the procedure is the same. The error message is included in a prompt to the LLM, together with similar information as in the prompt for formalizing goals; a list of objects in the scene, precondition alternatives, examples, a short scenario description, and specifications for the output. Just as with the LLM for goal interpretation, its reasoning can be extracted here as well. The main output is
\par\textbf{13. New precondition(s)}. The response from the \textit{LLM}, in the form of one or more preconditions to the failing action. These are then inserted into the tree as the first preconditions of the failing action and expanded further by the planner in order to ensure that it is satisfied before starting the action. 
\par The overall process can be repeated indefinitely if there is more preconditions missing for the action or if there are multiple actions missing preconditions. The result is a more complete BT policy that in the future handles similar issues automatically and retains the main properties of all BT policies, namely reactivity, modularity, and transparency.
\par We illustrate the failure resolution part of the method with an example shown in Fig.~\ref{fig:bt}. The example is the first task in Section~\ref{sec:experiments}. The instruction \textit{``Please put the blue cube on top of the green cube''} is interpreted by the LLM as the goal condition [``blue cube'' on ``green cube''] and expanded by the planner into the tree shown in Fig.~\ref{fig:bt}(a). When attempting the action [grasp ``blue cube''], an error occurs as a red cube is placed on top of the blue cube, blocking the path. The LLM correctly identifies [\textasciitilde ``any object'' on ``blue cube''?] as a necessary precondition for grasping the blue cube and the planner expands the precondition into the subtree on the bottom left of Fig.~\ref{fig:bt}(b), outlined in orange. Further, the already existing precondition [\textasciitilde grasped ``any object''?] now fails since the red cube is picked up after executing the subtree on the bottom left. The planner therefore expands it into the subtree at the bottom centre, outlined in blue.
\par We compare major properties of \ours~to some other methods in  Table~\ref{Method comparison}. Without formal long-horizon planning, LLMs tend to fail when attempting long-horizon tasks~\cite{valmeekam2022large}. Given the cost and time delay associated with each call to an LLM, a reactive policy is highly preferable, especially since it gives the possibility of verifying the policy before execution, otherwise every response from the LLM would have to be verified. Finally, no system will be perfect so failure resolution is vital. Assuming that any error that has happened once can happen again, a highly desirable trait and a main contribution of our method is for the system to learn from past failures so as not to repeat them. As far as we are aware, no other similar system fulfills all these requirements.

\section{Experiments and results}
We run several experiments to show that:
\begin{itemize}
    \item The changes to LLM-OBTEA~\cite{llmobtea} for goal interpretation are sound and give good results.
    \item \ours~can handle a variety of failures and identify both missing preconditions and parameters.
    \item The method is applicable in a real robot system on realistic tasks.
\end{itemize}

\begin{table}[htbp]
\vspace{.1cm}
\caption{Prompt results for varying levels of difficulty. All columns except the one titled ``Ours'' use LLM-OBTEAs original prompt. Methods denoted ``0F'' use no reflective feedback while ``5F'' means up to five rounds of reflective feedback. Ours is using GPT-4-1106 and no reflective feedback.  Our method achieves an almost perfect score while LLM-OBTEA struggles also with up to five rounds of reflective feedback.}
\begin{center}
\begin{tabular}{c c c c c}

\bf{Difficulty} & {\bf 0F GPT-3.5} & {\bf 5F GPT-3.5} & {\bf 0F GPT4 } & {\bf Ours}\cr
 
\cmidrule(r){1-1} \cmidrule(l){2-5}
Easy & 84.7\% & 90.7\% & 90.0\% & 100.0\% \cr
\cmidrule(lr){1-1} \cmidrule(lr){2-5}
Medium & 76.7\% & 82.0\% & 86.7\% & 100.0\% \cr
\cmidrule(lr){1-1} \cmidrule(lr){2-5}
Hard & 59.0\% & 65.0\% & 85.5\% & 97.0\% \cr
\bottomrule
\end{tabular}
\end{center}
\vspace{-.5cm}
\label{Prompt results}
\end{table}

\subsection{LLM-OBTEA comparisons}
The aim of these experiments is to compare the performance of our method for goal interpretation to LLM-OBTEA~\cite{llmobtea}. The main improvement is that we do not use the time and resource consuming reflective feedback. To enable this, we use an improved prompt and the more advanced LLM GPT-4-1106~\cite{gpt4techreport} as compared to LLM-OBTEA results obtained with GPT-3.5. We use the problem set from~\cite{llmobtea} with 100 different tasks in a cafe setting, grouped into three difficulty levels depending on the complexity of the logic of the task.
Table~\ref{Prompt results} shows the results for tasks with varying levels of difficulty. Our method achieves an almost perfect score while LLM-OBTEA struggles, requiring up to five rounds of reflective feedback. The results indicate that reflective feedback can be omitted and the LLM only needs to be queried once, saving time and cost. We also ran the original LLM-OBTEA prompt with GPT-4 to see how much of the improvement can be attributed to the improved LLM and it is clear that changing from GPT-3.5 to GPT-4 drastically improved the results, although it does not account for the whole difference.

To determine how much of the improvement can be attributed to various modifications of the prompt, we ran a number of ablations, as seen in Table~\ref{Ablation results}. The main improvements to the prompt that we ran ablations for are
\begin{itemize}
    \item Short descriptions of the conditions, instead of just condition names. For example \textit{``Active: The appliance is on. Negating turns the appliance off''}, avoids confusion with the condition \textit{On} meaning the object is placed on top of something.
    \item Strict specification that only the listed objects can be used and to use the most similar if necessary. This would otherwise be a problem for example if the user asks for ``fries'' but only ``chips'' are listed.
    \item Minor updates to the examples to make them more clear. Mainly, one example was ``turn up the air conditioning'' with the correct answer being to raise the temperature. However, most humans, as well as GPT-4, interpret ``turn up the air conditioning'' as ``increase the effect of the air conditioning, thereby lowering the temperature''. The effect is that GPT-4 gave the inverse answer to all instructions involving AC temperature. We added the word temperature to make the example less ambiguous.
\end{itemize}

\begin{table}[htbp]
\vspace{.1cm}
\caption{Prompt ablations. ``Ours'' is our complete improved prompt. ``No desc'' has no condition descriptions. ``No obj spec'' has no specification to only use listed objects. ``Orig ex'' has the original examples. ``CoT'' uses a chain of though prompt with reasoning before the answer.}
\begin{center}
\begin{tabular}{c c c c c c}

\bf{Difficulty} & {\bf Ours} & {\bf No desc.}& {\bf No obj spec.} & {\bf Orig ex} & {\bf CoT}  \cr
 
\cmidrule(r){1-1}\cmidrule(l){2-6}
Easy & 100.0\% & 100.0\% & 96.7\% & 91.3\% & 98.0\% \cr
\cmidrule(lr){1-1}\cmidrule(lr){2-6}
Medium & 100.0\% & 93.3\% & 90.0\% & 93.3\% & 96.0\% \cr
\cmidrule(lr){1-1}\cmidrule(lr){2-6}
Hard & 97.0\% & 88.5\% & 86.0\% & 91.0\% & 93.5\% \cr
\bottomrule
\end{tabular}
\end{center}
\vspace{-.5cm}
\label{Ablation results}
\end{table}
We also tested prompting the LLM to reason about the answer before giving the conditions, as chain of thought (CoT) prompting has led to good performance in previous works~\cite{wei2022chain}. We found that this instead lowered the accuracy. The tendency was that occasionally the LLM seemed to overthink the problem, making it more complicated than necessary. A simple example is a task where the given instruction is to bring a glass of water and the correct answer \textit{On\_Water\_Bar2}. This is perfectly handled without CoT, but with CoT the LLM reasons that it needs both a glass and water which is marked as wrong in the benchmark. Interestingly, we found that GPT-4 works sequentially so if the same prompt asks for the conditions first and reasoning after, the results are almost the same as without reasoning.
\par From Table~\ref{Ablation results} it is clear that the changes to the prompt had significant effects. Removing either one of the condition descriptions or object specification degraded the results on the hard problems almost to the level of the original prompt. It is also clear that all changes are needed as none could be removed without losing performance.

\begin{table*}[htbp]
\centering
\vspace{.1cm}
\caption{Ten example scenarios for identifying missing preconditions. Task descriptions are shortened to save space. We describe the cause of the problem in the middle column, this information is not given to the LLM. The text in the last column however is given to the LLM.}
\vspace{-.2cm}
\begin{center}
\begin{tabular}{c c c}

\fontsize{8}{0}\selectfont \thead{Task} & \fontsize{8}{0}\selectfont \thead{Problem and missing precondition} & \fontsize{8}{0}\selectfont \thead{Error message given} \cr
 
\cmidrule(lr){1-3}
Put the blue cube on the green cube & \makecell{A red cube is blocking the blue cube \\ and must be removed first} & No collision free path found \cr
\cmidrule(lr){1-3}
Put the blue cube on the green cube & \makecell{Two cubes are blocking both the blue and the green cube} & No collision free path found \cr
\cmidrule(lr){1-3}
Put the green cube in the red cup & The red cup is upside down and must be turned first & No collision free path found \cr
\cmidrule(lr){1-3}
Put the test tube in the centrifuge  & The centrifuge is closed and must be opened & No collision free path found \cr
\cmidrule(lr){1-3}
Put the plate in cupboard & The cupboard is locked and must be unlocked first & Torque limit exceeded \cr
\cmidrule(lr){1-3}
Bring me a banana & The bananas position is not known and it must be found first & Object "banana" is not in the dictionary \cr
\cmidrule(lr){1-3}
Bring me a banana & The banana is to far away, the robot must move closer first & Position of out reach \cr
\cmidrule(lr){1-3}
Bring coffee to table & There is no coffee yet, it must be made first & "coffee" not found \cr
\cmidrule(lr){1-3}
Bring fries or dessert & \makecell{The fries are at a different table \\ and the robot must move there first }& Position of out reach \cr
\cmidrule(lr){1-3}
Sweep the floor & The robot is not holding the mop so cleaning has no effect & \makecell{Postcondition\ IsClean\_Floor not met \\ after Sweep action completion} \cr

\bottomrule
\end{tabular}
\end{center}
\vspace{-.5cm}
\label{Precond results}
\end{table*}

\subsection{Identifying missing preconditions}
\label{sec:experiments}
In order to test the capability of our error resolution method, we tested it on a diverse set of 10 different scenarios. The task of the LLM is to identify which preconditions are missing for the failing action to successfully complete. Each scenario was run by the LLM 10 times to test its robustness. Due to space constraints, we only give short descriptions here but the full prompts are available in the code repository. The BT results for the first task can be seen in Fig.~\ref{fig:bt} and Table~\ref{Precond results} briefly describes all 10 scenarios. The last three scenarios are adapted from each of the difficulty levels of the LLM-OBTEA dataset.

All 10 scenarios were solved by our method with a perfect score but none of the scenarios can be solved without manual intervention using LLM-OBTEA or, to the best of our knowledge, any other method in a way that updates a transparent and reactive policy.

\subsection{Selecting missing parameters}
We also ran scenarios with missing parameters to show and exemplify the versatility of the approach. In some cases, the parameters of some behaviors can not be uniquely determined by the planner. Instead of requiring manual intervention to specify the values, we can utilize the LLM to suggest values for the parameters, based on the scenario and task description. We first asked the system to bring an egg and a hammer without specifying the grasp force. Over 10 runs, it suggested 5.3N on average for the egg and 37.2N for the hammer, both reasonable values. We then ran tests without specifying movement speed. For bringing a pillow it suggested 0.6m/s on average. When instead asked to ``bring a first aid kit so I can stop the bleeding'' it realized the urgency and suggested 1.5m/s. In addition, we asked it to put a baby in a crib. Understanding that the baby needs careful handling, the system suggested 0.1m/s. Lastly, we ran two scenarios with categorical parameters and asked it to specify a tool for putting sand in a bucket or cleaning a plate. The system sensibly suggested ``shovel'' for moving sand and ``sponge'' or ``brush'' for cleaning the plate. Note that the parameter values do not necessarily change the structure of the tree (although they could), but by intertwining with the planner, the value of the parameters automatically propagates with the planner to all relevant subtrees so that for example the shovel is used for all movements when handling sand.


\subsection{Real robot experiments}
We implemented and tested a complete solution on a real system to show that our method is sufficient for solving the tasks, given an adequate vision system, see Fig.~\ref{fig:centrifuge}. In these examples, the only input necessary from the user are short, natural language instructions like \textit{``Put the blue cube on the green cube''}. For simulating the system during the planning phase we use Open3D~\cite{open3d} for 3D computations, ignoring physics and instead teleporting the objects when necessary. We use an ABB YuMi robot with an Azure Kinect camera mounted on top with \textit{YoloWorld}~\cite{cheng2024yolo} for object detection and \textit{NanoSAM}~\cite{nanosam}, a distilled and much faster extension of \textit{MobileSAM}~\cite{mobile_sam}, for segmentation, and then use depth data with some simple heuristics to obtain the position estimates of the objects. Scene descriptions are retrieved by prompting GPT-4 with a camera image. We found that both YoloWorld and GPT-4 had problems separating stacked cubes and instead tended to identify one multicolored block, but if the prompt also included the task instruction it could correctly identify all cubes. The underlying BT framework used is \textit{PyTrees}~\cite{pytrees}, version 2.2.2. Specifically, we use a forked version with slightly changed visuals~\cite{pytreesfork}. For interfacing with the robot we use the ABBs \textit{RWS} API~\cite{rws}  and ABBs Automatic Path Planning algorithm for obtaining collision free paths.  In the example with voice commands, we use gTTS~\cite{gtts} to translate between text and speech. Recorded runs of tasks 1-4 can be seen in the accompanying video, showing that the method can be successfully applied to real systems and realistic tasks.

\section{Conclusions}
We present \ours, a method that takes natural language input and combines LLMs and long-horizon task planning to generate a reactive policy in the form of a Behavior Tree. We showed that with improved prompts and new LLMs, reflective feedback is not necessary and we can achieve high accuracy for goal interpretation by only prompting the LLM once per task, even for complex instructions. Further, we tested the failure resolution capabilities of the method on a variety of tasks for identifying missing preconditions or parameters and showed that is capable of reliably solving diverse problems. Finally we implemented our method on an ABB YuMi system for a subset of the tasks and successfully executed them to show our method's validity in a real setting.

\section{Future work}
There are still some questions that remain unanswered in this paper and that had to be left to future work. For instance, the tests performed are somewhat limited with at most a few dozen available objects and conditions. It is not clear where the limit for the planner and LLM lies when the number of objects and conditions increase to thousands or more. It would also be interesting to study whether the combination with the planner can be used to resolve ambiguous instructions without extended communication with the user, by for example ruling out branches that the planner deems unsolvable. Another case we did not study is when the skill library is missing the necessary actions to solve the task. Utilizing the LLM to create those actions from lower level primitives is an interesting prospect.

\clearpage
\bibliographystyle{IEEEtran}
\bibliography{IEEEabrv, biblio, references, llmbiblio}

\begin{thebibliography}{10}
\providecommand{\url}[1]{#1}
\csname url@rmstyle\endcsname
\providecommand{\newblock}{\relax}
\providecommand{\bibinfo}[2]{#2}
\providecommand\BIBentrySTDinterwordspacing{\spaceskip=0pt\relax}
\providecommand\BIBentryALTinterwordstretchfactor{4}
\providecommand\BIBentryALTinterwordspacing{\spaceskip=\fontdimen2\font plus
\BIBentryALTinterwordstretchfactor\fontdimen3\font minus \fontdimen4\font\relax}
\providecommand\BIBforeignlanguage[2]{{%
\expandafter\ifx\csname l@#1\endcsname\relax
\typeout{** WARNING: IEEEtran.bst: No hyphenation pattern has been}%
\typeout{** loaded for the language `#1'. Using the pattern for}%
\typeout{** the default language instead.}%
\else
\language=\csname l@#1\endcsname
\fi
#2}}

\bibitem{worldrobotics}
C.~Müller, W.~Kraus, B.~Graf, and K.~E. Bregler, ``World robotics 2023 – service robots,'' IFR Statistical Department, Tech. Rep., 2023.

\bibitem{iovino_survey_2022}
M.~Iovino, E.~Scukins, J.~Styrud, P.~{\"O}gren, and C.~Smith, ``A survey of {{Behavior Trees}} in robotics and {{AI}},'' \emph{Robotics and Autonomous Systems}, vol. 154, p. 104096, Aug. 2022.

\bibitem{colledanchise_behavior_2018}
M.~Colledanchise and P.~{\"O}gren, \emph{Behavior {{Trees}} in {{Robotics}} and {{AI}} : {{An Introduction}}}.\hskip 1em plus 0.5em minus 0.4em\relax {CRC Press}, July 2018.

\bibitem{colledanchise_synthesis_2017}
M.~Colledanchise, R.~M. Murray, and P.~Ögren, ``Synthesis of correct-by-construction behavior trees,'' in \emph{2017 {IEEE}/{RSJ} {International} {Conference} on {Intelligent} {Robots} and {Systems} ({IROS})}, Sept. 2017, pp. 6039--6046.

\bibitem{colledanchise_learning_2019}
M.~Colledanchise, R.~Parasuraman, and P.~{\"O}gren, ``Learning of {{Behavior Trees}} for {{Autonomous Agents}},'' \emph{IEEE Transactions on Games}, vol.~11, no.~2, pp. 183--189, June 2019.

\bibitem{colledanchise_towards_2019}
M.~Colledanchise, D.~Almeida, and P.~{\"O}gren, ``Towards {{Blended Reactive Planning}} and {{Acting}} using {{Behavior Trees}},'' in \emph{2019 {{International Conference}} on {{Robotics}} and {{Automation}} ({{ICRA}})}, May 2019, pp. 8839--8845.

\bibitem{rovida_extended_2017}
F.~Rovida, B.~Grossmann, and V.~Krüger, ``Extended behavior trees for quick definition of flexible robotic tasks,'' in \emph{2017 {IEEE}/{RSJ} {International} {Conference} on {Intelligent} {Robots} and {Systems} ({IROS})}, Sept. 2017, pp. 6793--6800.

\bibitem{iovino_learning_2021}
M.~Iovino, J.~Styrud, P.~Falco, and C.~Smith, ``Learning {{Behavior Trees}} with {{Genetic Programming}} in {{Unpredictable Environments}},'' in \emph{2021 {{IEEE International Conference}} on {{Robotics}} and {{Automation}} ({{ICRA}})}, May 2021, pp. 4591--4597.

\bibitem{styrud_combining_2022}
J.~Styrud, M.~Iovino, M.~Norrl{\"o}f, M.~Bj{\"o}rkman, and C.~Smith, ``Combining {{Planning}} and {{Learning}} of {{Behavior Trees}} for {{Robotic Assembly}},'' in \emph{2022 {{International Conference}} on {{Robotics}} and {{Automation}} ({{ICRA}})}, May 2022, pp. 11\,511--11\,517.

\bibitem{llmbt}
H.~Zhou, Y.~Lin, L.~Yan, J.~Zhu, and H.~Min, ``Llm-bt: Performing robotic adaptive tasks based on large language models and behavior trees,'' \emph{arXiv preprint arXiv:2404.05134}, 2024.

\bibitem{llmobtea}
X.~Chen, \emph{et~al.}, ``Integrating intent understanding and optimal behavior planning for behavior tree generation from human instructions,'' \emph{arXiv preprint arXiv:2405.07474}, 2024.

\bibitem{radford2019language}
A.~Radford, \emph{et~al.}, ``Language models are unsupervised multitask learners,'' \emph{OpenAI blog}, vol.~1, no.~8, p.~9, 2019.

\bibitem{brown2020language}
T.~B. Brown, ``Language models are few-shot learners,'' \emph{arXiv preprint arXiv:2005.14165}, 2020.

\bibitem{wang2024survey}
L.~Wang, \emph{et~al.}, ``A survey on large language model based autonomous agents,'' \emph{Frontiers of Computer Science}, vol.~18, no.~6, p. 186345, 2024.

\bibitem{saycan}
A.~Brohan, \emph{et~al.}, ``Do as i can, not as i say: Grounding language in robotic affordances,'' in \emph{Conference on robot learning}.\hskip 1em plus 0.5em minus 0.4em\relax PMLR, 2023, pp. 287--318.

\bibitem{rana2023sayplan}
K.~Rana, J.~Haviland, S.~Garg, J.~Abou-Chakra, I.~Reid, and N.~Suenderhauf, ``Sayplan: Grounding large language models using 3d scene graphs for scalable task planning,'' \emph{arXiv preprint arXiv:2307.06135}, 2023.

\bibitem{hazra2024saycanpay}
R.~Hazra, P.~Z. Dos~Martires, and L.~De~Raedt, ``Saycanpay: Heuristic planning with large language models using learnable domain knowledge,'' in \emph{Proceedings of the AAAI Conference on Artificial Intelligence}, vol.~38, no.~18, 2024, pp. 20\,123--20\,133.

\bibitem{yang2024text2reaction}
Z.~Yang, \emph{et~al.}, ``Text2reaction: Enabling reactive task planning using large language models,'' \emph{IEEE Robotics and Automation Letters}, 2024.

\bibitem{aeronautiques1998pddl}
C.~Aeronautiques, \emph{et~al.}, ``Pddl| the planning domain definition language,'' \emph{Technical Report, Tech. Rep.}, 1998.

\bibitem{liu2023llm+}
B.~Liu, \emph{et~al.}, ``Llm+ p: Empowering large language models with optimal planning proficiency,'' \emph{arXiv preprint arXiv:2304.11477}, 2023.

\bibitem{hallen2024behavior}
M.~Hallen, M.~Iovino, S.~Sander-Tavallaey, and C.~Smith, ``Behavior trees in industrial applications: A case study in underground explosive charging,'' in \emph{2024 IEEE 20th International Conference on Automation Science and Engineering (CASE)}.\hskip 1em plus 0.5em minus 0.4em\relax IEEE, 2024.

\bibitem{iovino2023framework}
M.~Iovino, J.~Styrud, P.~Falco, and C.~Smith, ``A framework for learning behavior trees in collaborative robotic applications,'' \emph{2023 IEEE International Conference on Automation Science and Engineering (CASE)}, 2023.

\bibitem{mayr22priors}
M.~Mayr, C.~Hvarfner, K.~Chatzilygeroudis, L.~Nardi, and V.~Krueger, ``Learning skill-based industrial robot tasks with user priors,'' in \emph{2022 IEEE 18th International Conference on Automation Science and Engineering (CASE)}.\hskip 1em plus 0.5em minus 0.4em\relax IEEE, 2022, pp. 1485--1492.

\bibitem{tumova_maximally_2014}
\BIBentryALTinterwordspacing
J.~Tumova, A.~Marzinotto, D.~V. Dimarogonas, and D.~Kragic, ``\BIBforeignlanguage{en}{Maximally satisfying {LTL} action planning},'' in \emph{\BIBforeignlanguage{en}{2014 {IEEE}/{RSJ} {International} {Conference} on {Intelligent} {Robots} and {Systems}}}.\hskip 1em plus 0.5em minus 0.4em\relax Chicago, IL, USA: IEEE, Sept. 2014, pp. 1503--1510. [Online]. Available: \url{http://ieeexplore.ieee.org/document/6942755/}
\BIBentrySTDinterwordspacing

\bibitem{holzl_reasoning_2015}
M.~Hölzl and T.~Gabor, ``\BIBforeignlanguage{en}{Reasoning and {Learning} for {Awareness} and {Adaptation}},'' in \emph{\BIBforeignlanguage{en}{Software {Engineering} for {Collective} {Autonomic} {Systems}: {The} {ASCENS} {Approach}}}, ser. Lecture {Notes} in {Computer} {Science}, M.~Wirsing, M.~Hölzl, N.~Koch, and P.~Mayer, Eds.\hskip 1em plus 0.5em minus 0.4em\relax Cham: Springer International Publishing, 2015, pp. 249--290.

\bibitem{gustavsson_combining_2022}
O.~Gustavsson, M.~Iovino, J.~Styrud, and C.~Smith, ``Combining {{Context Awareness}} and {{Planning}} to {{Learn Behavior Trees}} from {{Demonstration}},'' in \emph{2022 31st {{IEEE International Conference}} on {{Robot}} and {{Human Interactive Communication}} ({{RO-MAN}})}, Aug. 2022, pp. 1153--1160.

\bibitem{gugliermo2023learning}
S.~Gugliermo, E.~Schaffernicht, C.~Koniaris, and F.~Pecora, ``Learning behavior trees from planning experts using decision tree and logic factorization,'' \emph{IEEE Robotics and Automation Letters}, vol.~8, no.~6, pp. 3534--3541, 2023.

\bibitem{iovino2022interactive}
M.~Iovino, F.~I. Do{\u{g}}an, I.~Leite, and C.~Smith, ``Interactive disambiguation for behavior tree execution,'' in \emph{2022 IEEE-RAS 21st International Conference on Humanoid Robots (Humanoids)}.\hskip 1em plus 0.5em minus 0.4em\relax IEEE, 2022, pp. 82--89.

\bibitem{mayr2022combining}
M.~Mayr, F.~Ahmad, K.~Chatzilygeroudis, L.~Nardi, and V.~Krueger, ``Combining planning, reasoning and reinforcement learning to solve industrial robot tasks,'' \emph{IROS 2022 Workshop on Trends and Advances in Machine Learning and Automated Reasoning for Intelligent Robots and Systems}, 2022.

\bibitem{mayr2022skill}
------, ``Skill-based multi-objective reinforcement learning of industrial robot tasks with planning and knowledge integration,'' in \emph{2022 IEEE International Conference on Robotics and Biomimetics (ROBIO)}.\hskip 1em plus 0.5em minus 0.4em\relax IEEE, 2022, pp. 1995--2002.

\bibitem{styrud2024bebop}
J.~Styrud, M.~Mayr, E.~Hellsten, V.~Krueger, and C.~Smith, ``Bebop-combining reactive planning and bayesian optimization to solve robotic manipulation tasks,'' in \emph{2024 IEEE International Conference on Robotics and Automation (ICRA)}.\hskip 1em plus 0.5em minus 0.4em\relax IEEE, 2024, pp. 16\,459--16\,466.

\bibitem{izzo2024btgenbot}
R.~A. Izzo, G.~Bardaro, and M.~Matteucci, ``Btgenbot: Behavior tree generation for robotic tasks with lightweight llms,'' \emph{arXiv preprint arXiv:2403.12761}, 2024.

\bibitem{mower2024ros}
C.~E. Mower, \emph{et~al.}, ``Ros-llm: A ros framework for embodied ai with task feedback and structured reasoning,'' \emph{arXiv preprint arXiv:2406.19741}, 2024.

\bibitem{chen2024efficient}
X.~Chen, \emph{et~al.}, ``Efficient behavior tree planning with commonsense pruning and heuristic,'' \emph{arXiv preprint arXiv:2406.00965}, 2024.

\bibitem{wang2024mosaic}
H.~Wang, \emph{et~al.}, ``Mosaic: A modular system for assistive and interactive cooking,'' \emph{arXiv preprint arXiv:2402.18796}, 2024.

\bibitem{tagliamonte2024generalizable}
C.~Tagliamonte, D.~Maccaline, G.~LeMasurier, and H.~A. Yanco, ``A generalizable architecture for explaining robot failures using behavior trees and large language models,'' in \emph{Companion of the 2024 ACM/IEEE International Conference on Human-Robot Interaction}, 2024, pp. 1038--1042.

\bibitem{gao2024physically}
J.~Gao, \emph{et~al.}, ``Physically grounded vision-language models for robotic manipulation,'' in \emph{2024 IEEE International Conference on Robotics and Automation (ICRA)}.\hskip 1em plus 0.5em minus 0.4em\relax IEEE, 2024, pp. 12\,462--12\,469.

\bibitem{yang2018learning}
C.~Yang, C.~Zeng, Y.~Cong, N.~Wang, and M.~Wang, ``A learning framework of adaptive manipulative skills from human to robot,'' \emph{IEEE Transactions on Industrial Informatics}, vol.~15, no.~2, pp. 1153--1161, 2018.

\bibitem{li2019robot}
F.~Li, Q.~Jiang, S.~Zhang, M.~Wei, and R.~Song, ``Robot skill acquisition in assembly process using deep reinforcement learning,'' \emph{Neurocomputing}, vol. 345, pp. 92--102, 2019.

\bibitem{biggar2020framework}
O.~Biggar and M.~Zamani, ``A framework for formal verification of behavior trees with linear temporal logic,'' \emph{IEEE Robotics and Automation Letters}, vol.~5, no.~2, pp. 2341--2348, 2020.

\bibitem{valmeekam2022large}
K.~Valmeekam, A.~Olmo, S.~Sreedharan, and S.~Kambhampati, ``Large language models still can't plan (a benchmark for llms on planning and reasoning about change),'' in \emph{NeurIPS 2022 Foundation Models for Decision Making Workshop}, 2022.

\bibitem{gpt4techreport}
J.~Achiam, \emph{et~al.}, ``Gpt-4 technical report,'' \emph{arXiv preprint arXiv:2303.08774}, 2023.

\bibitem{wei2022chain}
J.~Wei, \emph{et~al.}, ``Chain-of-thought prompting elicits reasoning in large language models,'' \emph{Advances in neural information processing systems}, vol.~35, pp. 24\,824--24\,837, 2022.

\bibitem{open3d}
Q.-Y. Zhou, J.~Park, and V.~Koltun, ``{Open3D}: {A} modern library for {3D} data processing,'' \emph{arXiv:1801.09847}, 2018.

\bibitem{cheng2024yolo}
T.~Cheng, L.~Song, Y.~Ge, W.~Liu, X.~Wang, and Y.~Shan, ``Yolo-world: Real-time open-vocabulary object detection,'' in \emph{Proceedings of the IEEE/CVF Conference on Computer Vision and Pattern Recognition}, 2024, pp. 16\,901--16\,911.

\bibitem{nanosam}
\BIBentryALTinterwordspacing
``Nanosam.'' [Online]. Available: \url{https://github.com/NVIDIA-AI-IOT/nanosam}
\BIBentrySTDinterwordspacing

\bibitem{mobile_sam}
C.~Zhang, \emph{et~al.}, ``Faster segment anything: Towards lightweight sam for mobile applications,'' \emph{arXiv preprint arXiv:2306.14289}, 2023.

\bibitem{pytrees}
\BIBentryALTinterwordspacing
D.~Stonier, ``Pytrees.'' [Online]. Available: \url{https://github.com/splintered-reality/py_trees}
\BIBentrySTDinterwordspacing

\bibitem{pytreesfork}
\BIBentryALTinterwordspacing
``Pytrees fork.'' [Online]. Available: \url{https://github.com/jstyrud/py_trees}
\BIBentrySTDinterwordspacing

\bibitem{rws}
\BIBentryALTinterwordspacing
``Rws.'' [Online]. Available: \url{https://developercenter.robotstudio.com/api/RWS}
\BIBentrySTDinterwordspacing

\bibitem{gtts}
\BIBentryALTinterwordspacing
P.~N. Durette, ``gtts.'' [Online]. Available: \url{https://github.com/pndurette/gTTS}
\BIBentrySTDinterwordspacing

\end{thebibliography}

\end{document}